\documentclass[11pt,a4paper]{article}
\usepackage[hyperref]{acl2019}
\usepackage{times}
\usepackage{latexsym}
\usepackage{booktabs}
\usepackage{times}
\usepackage{latexsym}
\usepackage{graphicx}
\usepackage{algorithm}
\usepackage{amsfonts}
\usepackage{amssymb}
\usepackage{listings}
\usepackage{amsmath}
\usepackage{float}
\usepackage{comment}
\usepackage{algpseudocode}
\usepackage{xspace}
\usepackage{multirow}
\usepackage{url}
\usepackage{tikz}
\usepackage{pgfplots}
\usepackage{todonotes}
\pgfplotsset{width=7cm,compat=1.3}
\usepackage{filecontents}
\usetikzlibrary{shapes, positioning, decorations.pathreplacing, shapes.multipart,arrows,chains,calc,plotmarks}

\usepackage{subfig}
\allowdisplaybreaks

\aclfinalcopy % Uncomment this line for the final submission
\title{BP-Transformer: Modelling Long-Range Context via Binary Partitioning}

\author{Zihao Ye\footnotemark[2] , Qipeng Guo\footnotemark[2] \hspace{0.5mm}\footnotemark[3] \hspace{0.5mm}\thanks{Work done during internship at AWS Shanghai AI Lab.} , Quan Gan\footnotemark[2] ,  Xipeng Qiu\footnotemark[3] \hspace{0.5mm}, Zheng Zhang\footnotemark[2] \hspace{0.5mm}\footnotemark[4] \\
\footnotemark[2]\hspace{0.5mm} AWS Shanghai AI Lab\\
\footnotemark[3]\hspace{0.5mm} Fudan University\\
\footnotemark[4]\hspace{0.5mm} New York University Shanghai \\
\texttt{\{yeziha, gqipeng, quagan, zhaz\}@amazon.com, xpqiu@fudan.edu.cn}
}

\date{}

\begin{document}

\maketitle

\begin{abstract}
The Transformer model is widely successful on many natural language processing tasks. However, the quadratic complexity of self-attention limit its application on long text.
%Transformer-XL explores segment-level recurrence mechanism to capture long-term dependency, but its recurrent nature rendering it unable to model bidirectional dependency among words.
In this paper, adopting a fine-to-coarse attention mechanism on multi-scale spans via binary partitioning (BP), we propose BP-Transformer (BPT for short). BPT yields $O(k\cdot n\log (n/k))$ connections where $k$ is a hyperparameter to control the density of attention. BPT has a good balance between computation complexity and model capacity. A series of experiments on text classification, machine translation and language modeling shows BPT has a superior performance for long text than previous self-attention models.
Our code, hyperparameters and CUDA kernels for sparse attention are available in PyTorch \footnote{\url{https://github.com/yzh119/BPT}}.
\end{abstract}

\section{Introduction}

Transformer, a self-attention based model, has achieved many impressive results on Natural Language Processing (NLP) tasks, notably machine translation \citep{vaswani2017attention}, language modeling \citep{radford2018improving}, and text classification \citep{devlin2018bert}.
However, its self-attention mechanism imposes a quadratic cost with respect to sequence length, limiting its wider application, especially for long text.

%\citet{al2018character} has shown the effectiveness of introducing longer context in Transformer-based models.
To address this problem, some previous works have explored different directions. (1) Hierarchical Transformers~\citep{miculicich-etal-2018-document,liu2019hierarchical} uses two Transformers in a hierarchical architecture: one Transformer models the sentence representation with word-level context, and another the document representation with the sentence-level context. %However, hierarchical structures is simple and hard to refine the word representation with long-range context.
(2) Lightweight Transformers~\citep{child2019sparsetransformer,sukhbaatar2019adaptive,guo2019startransformer,dai2019transformer} reduce the complexity by reconstructing the connections between tokens.

%For example, Transformer-XL~\citep{dai2019transformer} divides sequence into segments, each word in a segment attends to the current and previous segment, forming in a recurrent connection between the segments. The recurrence nature of Transformer-XL renders it unable to model bidirectional dependency among words in a layer. Transformer-XL also uses stop-gradient mechanism between segments, impeding supervision signal from later segments to be back-propagated to earlier ones.

\begin{figure}
\centering
\includegraphics[width=0.48\textwidth]{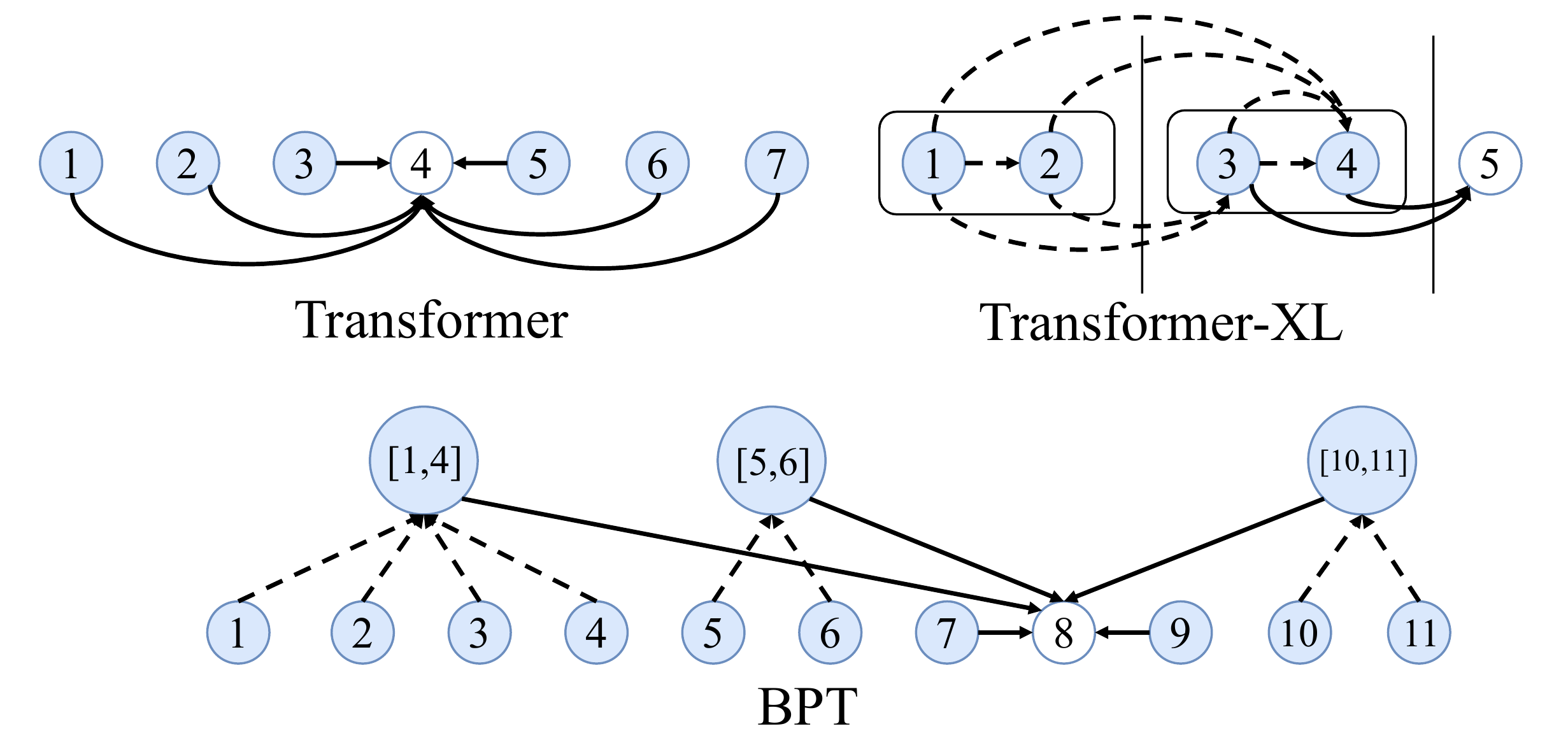}
\caption{Different attention pattern in Transformer-like models. Solid line refers to direct attention, while the dashed line denotes dependency. Unrelated connections and self loops are omitted for clarity.}
\label{fig:bpt1}
\end{figure}

Besides the computational cost, the fully-connected nature of Transformer does not incorporate the commonsensible inductive bias of language, such as sequential or syntax structure. The dependency relations between tokens are totally learned from scratch. Therefore, Transformer usually performs better on huge datasets and is easy to overfit on small datasets~\citep{guo2019startransformer}.

The above observation motivates us to explore better structure for self-attention models to balance the capability and computation complexity. In this paper, we propose a new architecture called \textbf{BP-Transformer} (BPT for short), which partitions the input sequence into different multi-scale spans via \textbf{binary partitioning} (BP). BPT incorporates an inductive bias of attending the context information from fine-grain to coarse-grain as the relative distance increases. The farther the context information is, the coarser its representation is.
%We can use a binary partition tree to arrange the different spans in a hierarchical structure.
BPT can be regard as graph neural network, whose nodes are the multi-scale spans.
%The edge construction is based on the following ideas. The representation of each span is computed  directly from its containing leaf nodes.
A token node can attend the smaller-scale span for the closer context and the larger-scale span for the longer-distance context.
The representations of nodes are updated with Graph Self-Attention~\citep{velickovic2017graph}.

%over all words mapped to its descendent leaves: nodes at level $l$ summarizes a range of size $2^l$. The idea stems from \emph{Segment Tree} data structure \citep{de1997computational}, where a set of internal nodes are created storing interval information.

%Vanilla Transformer uses ``positional encoding'' to inject absolute position information of each token in the sequence. \citep{shaw2018self} show that instead of modeling absolute position directly, relative position better encodes position information in Transformers.

Moreover, to better represent the position information of the span nodes and token nodes, we generalize the notion of relative position~\citep{shaw2018self} from sequences to trees and show that it better captures position bias.

Thus, BPT incorporates the advantages of both hierarchical and lightweight Transformerss: (1) it models the long-range context in an hierarchical fashion, (2) reduces computation cost with fewer edges, and finally, (3) introduces coarse-to-fine connections to approximate the reasonable inductive bias of language, with a net effect of making BPT easier to train.

We evaluate BPT on a variety of Sentence-Level and Document-Level NLP tasks: language modeling, machine translation and text classification. The experiment results show that BPT consistently outperforms previous self-attention based models. We also show that the inductive bias of BPT works nicely on short text and can scale to large datasets. Finally, we show BPT is faster and more memory efficient than vanilla Transformer when dealing with long sequence.

\section{Related Work}

\subsection{Recap: Transformer}
Given a sentence with $n$ input tokens, the Transformer model iteratively computes at layer $t$ the $d$-dimensional representations of each input token $\mathbf{H}^t \in \mathbb{R}^{n \times d}$, where $\mathbf{H}^0$ represents the initial token embeddings.
The core of a Transformer step is Multi-head Self-Attention (MSA), which can be formulated as follows:
\begin{equation}
\begin{gathered}
    \text{MSA}(\mathbf{H}) = [\text{head}_1,\cdots,\text{head}_h] W^O, \\ \text{head}_i = \text{softmax}\left(\frac{\mathbf{Q}_i \mathbf{K}_i^T}{\sqrt{d}}\right)\mathbf{V}_i, \\
    \mathbf{Q}_i = \mathbf{H} \mathbf{W}_i^Q, \quad % (L, q) = (L, d) * (d, q)
    \mathbf{K}_i = \mathbf{H} \mathbf{W}_i^K, \quad % (L, q) = (L, d) * (d, q)
    \mathbf{V}_i = \mathbf{H} \mathbf{W}_i^V, % (L, v) = (L, d) * (d, v)
\end{gathered}
\end{equation}
where $h$ is the number of heads, and $\mathbf{W}_i^Q$, $\mathbf{W}_i^K$, $\mathbf{W}_i^V$, $\mathbf{W}^O$ are learnable parameters.

Transformer then computes $\mathbf{H}^{t+1}$ from $\mathbf{H}^t$:
\begin{align}
    \mathbf{Z}^t = \text{norm}(\mathbf{H}^t + \text{MSA}(\mathbf{H}^t)), \\
     \mathbf{H}^{t+1} = \text{norm}(\mathbf{Z}^t + \text{FFN}(\mathbf{Z}^t)),
\end{align}
where $\text{norm}$ represents the layer normalization \citep{ba2016layer} and $\text{FFN}$ stands for the Position-wise Feed-Forward Network in \citep{vaswani2017attention}.
Note that each step $t$ has its own parameters.

\subsection{Hierarchical Attention}

Some previous work has explored the direction of applying self-attention on hierarchical features: HAN \citep{yang2016hierarchical} exploits a two-level attention mechanism that first applies self-attention on word features to get a sentence representation, then uses self-attention on sentence level features to get a document level features. \citet{shen2018bi} proposed a network structured called ``bi-directional block self-attention network(Bi-BloSAN)'' that divides a sequence into blocks, and sequentially applies intra-block attention and inter-block attention inside a layer. \citet{miculicich-etal-2018-document} uses a HAN structure to get sentence-level feature in Transformers for Document-Level Machine Translation. Different from them, our model updates hierarchical features synchronously inside a layer, and update them iteratively by stacking layers.

%: a two-level attention mechanism that first applies self-attention on word features to get a sentence representation, then uses self-attention on sentence level features to get a document level features .

\subsection{Lightweight Self-Attention }
Recently there has also been several works focusing on reducing the computational cost of Self-Attention in Transformers: T-DMCA \citep{LiuSPGSKS18} reduced the memory usage by first dividing the sequence tokens into blocks with similar length and performing attention inside each block independently. Sparse Transformer \citep{child2019sparsetransformer} decomposes attention into two categories: for a sequence with length $n$, we divide it into $\sqrt{n}$ equal-sized blocks. Each token attends to its previous tokens inside a $\sqrt{n}$ block it lies in, and to $\sqrt{n}$ previous blocks. Compared to our model, the Sparse Transformer does not maintain the representations of hierarchical features, and the computational cost of Sparse Transformer is $O(n\sqrt{n})$ while ours is $O(n\log n)$.
 Transformer-XL \citep{dai2019transformer} introduces the notion of recurrence into Transformer. It divides the input sequence into multiple segments and recurrently attends to the hidden states of the previous segments. They achieved state-of-the-art on several language modeling benchmarks. Compared to our model, Transformer-XL could only model sequences in one direction, making it hard to deal with tasks where bi-directional information is required. \citet{sukhbaatar2019adaptive} proposed a adaptive mechanism to learn optimal context length in transformers for each head per layer, thus reducing the total computational and memory cost of transformers. \citet{guo2019startransformer} suggest that the fully-connected nature of self-attention in Transformer is not a good inductive bias, they proposed Star-Transformer which links adjacency words coupled with a central relay node to capture both local and global dependencies, with such reduction, Star-Transformer achieved significant improvements against standard Transformer on moderate sized datasets. However, Star-Transformer is not suitable for auto-regressive models in which each word should only be conditioned on its previous words, while the relay node in Star-Transformer summarizes the whole sequence.

%On the other hand, the successes of Transformer come at the cost of large amount of computation and can only be obtained with huge datasets.

\section{Proposed Model}

In this paper, we balance the model capability and computation complexity by incorporating the inductive bias. The key insight is that not every token needs to be attended to for context representation. Instead, for an given input token, we can group its context into different-scale non-overlapping spans, and the scale of a span increases with its relative distance. That is, instead attending to every token, the input token attends to different spans away from it in a fine-to-coarse fashion.

We now describe our model as graph neural network and detail it in the following sections.

\subsection{Transformer as Graph Neural Networks}
\label{sec:trans-gnn}

A valid perspective is to view information fusing with self-attention in Transformer as message passing on a fully-connected graph, with input tokens as nodes and attentions between nodes as edges \cite{battaglia2018relational}. In particular, such a process is very similar to Graph Attention Network~\cite{velickovic2017graph}. Thus, different graph structure encodes different inductive bias of attention and results in different time/space complexity.

To describe Transformer in GNN framework, we first construct a fully-connected graph $\mathcal{G}$, in which each node is a token of the input sequence.  All nodes in $\mathcal{G}$ are interconnected and each node has a self-loop edge.

We extend the self-attention mechanism of Transformer to graph, called \textbf{Graph Self-Attention} (GSA). For a given node $u$, we update its representation according to its neighbour nodes, formulated as $\mathbf{h}^u \leftarrow \mathrm{GSA}(\mathcal{G}, \mathbf{h}^u)$.

Let $\mathcal{A}(u)$ denote the set of the neighbour nodes of $u$ in $\mathcal{G}$, $\mathrm{GSA}(\mathcal{G}, \mathbf{h}^u)$ is detailed as follows:
\begin{gather}\small
    \mathbf{A}^u = \text{concat}(\{\mathbf{h}_v \mid v \in \mathcal{A}(u)\}), \\
    \mathbf{Q}^u_i = \mathbf{H}_k \mathbf{W}_i^Q,\!  % (q,)
    \mathbf{K}^u_i = \mathbf{A}^u \mathbf{W}_i^K,\!  % (A, q)
    \mathbf{V}^u_i = \mathbf{A}^u \mathbf{W}_i^V,\! \\ % (A, v)
    \mathrm{head}_i^u = \text{softmax}\left(\dfrac{\mathbf{Q}^u_i {\mathbf{K}^u_i}^T}{\sqrt{d}}\right) \mathbf{V}^u_i, \label{eqn:no-relpos}\\
    \text{GSA}(\mathcal{G}, \mathbf{h}_u) = [\mathrm{head}^u_1,\cdots,\mathrm{head}^u_h] \mathbf{W}^O,
\end{gather}
where $d$ is the dimension of $\mathbf{\mathbf{h}}$, and $\mathbf{W}^Q_i, \mathbf{W}^K_i, \mathbf{W}^V_i$ are trainable parameters of the $i$-th attention head.

\subsection{Graph Construction}

%Instead of attending over every single token, BPT builds a graph that organize tokens into different partitions.

%Each token attends over either spans or other tokens, thereby reducing the number of nodes to attend.

\subsubsection{Node Construction}

To achieve the effect of fine-to-coarse attention, we partition a sequence into multi-granular spans via binary partitioning (BP).

Binary partitioning is a generic process of recursively dividing a sequence into two until the partitioning satisfies one or more requirements. In this paper, we use a simple rule to stop subdividing when a partition just contains a single token.
For a sequence with length $n$, there are $2n-1$ partitions. Figure \ref{fig:bpt1} illustrates the process of binary partitioning over a sequence.  Each partition can be regarded as a node in GNN and its representation is computed according to its contained tokens.

%, instead of attending $1,2,6,7$ directly, $4$ attends two summarization nodes $[1,2]$ and $[6,7]$ as they are  ``distant'' neighbors.

\begin{figure}[!htb]
\centering
\includegraphics[width=0.48\textwidth]{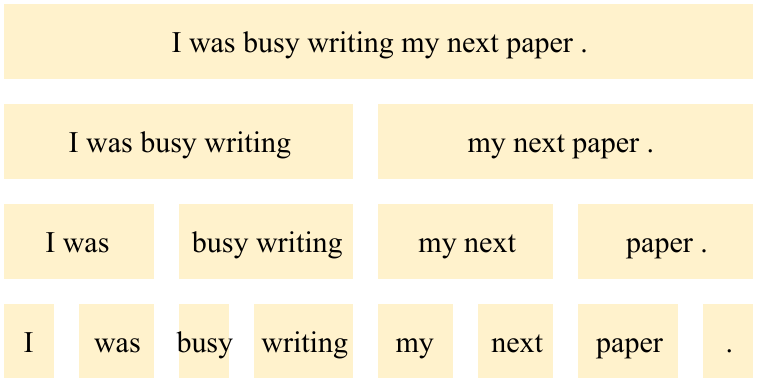}
\caption{Binary partitioning of a sequence.}
\label{fig:bpt3}
\end{figure}

% We note that generalizing the idea to arbitrary $k$-nary tree is possible, as long as the invariants (stated below) is abided to.

 \begin{figure*}[!htb]
\centering
\includegraphics[width=\textwidth]{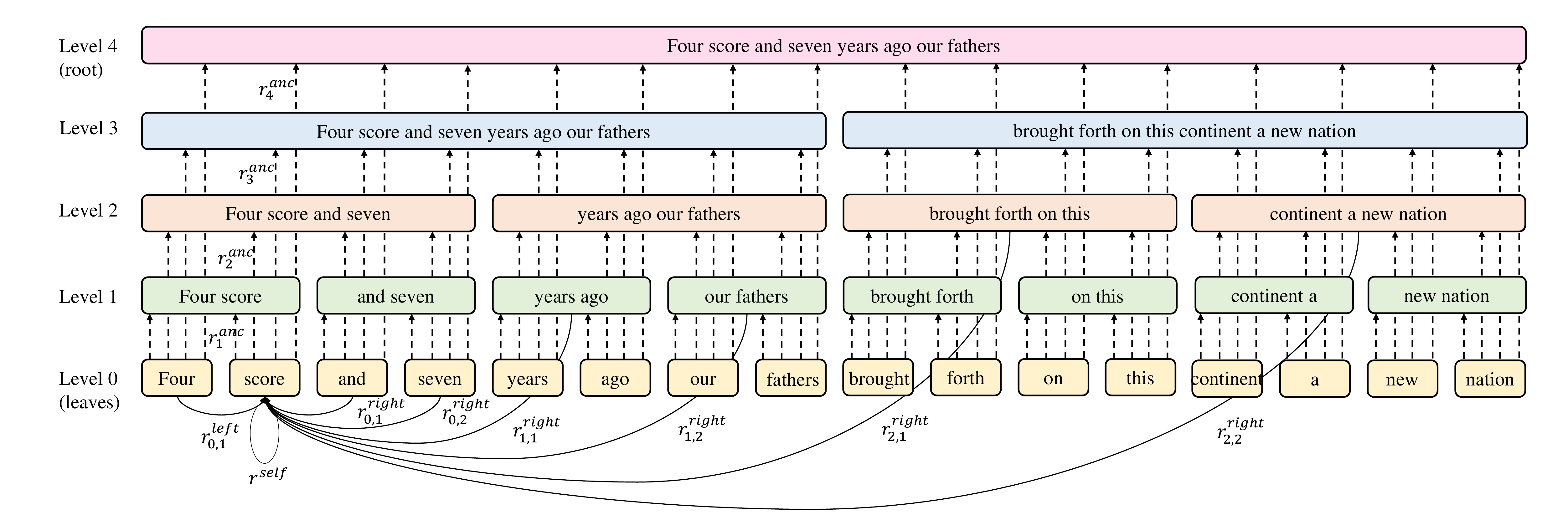}
\caption{The figure illustrates how to build the graph: nodes at different levels are colored differently, dashed lines are edges connects token nodes to span nodes; solid lines are edges connect to token nodes. The $r_{*}^{*}$ are relative positions assigned to edges.}
\label{fig:tree}
%\todo[inline]{fig3: modify the annotations $r^{\text{self}}$, $r^{\text{left}}_{j,i}$, $r^{\text{right}}_{j,i}$ and $r^{\text{anc}}_j$.}
\end{figure*}

The binary partitioning of a sequence constructs a perfect binary tree in which all internal nodes have two children and all leaf nodes have the same depth. Each leaf node corresponds to an input token in the sequence.

We simply divide the nodes into two types, \emph{token} and \emph{span}, both of which are used as nodes in our GNN construction:
\begin{description}
    \item [Token nodes] the leaf nodes in the binary partition tree.
    \item [Span nodes] the internal node of the tree, each has at least two child nodes.
\end{description}
%\paragraph{Token Nodes} The token node is the leaf node in the binary partition tree.

%\paragraph{Span Nodes} The span node is the internal node the binary partition tree, which contains at least two tokens.

%All the token nodes and span nodes are used as the nodes in our GNN.

\subsubsection{Edge Construction}

The binary partitioning generating a binary tree. For a sequence with $n$ tokens, we have $n$ token nodes and $n-1$ span nodes. Formally, let $u_{l,m}$ denote the $m$-th node at level $l$. The level increases in bottom-up fashion. The level of token nodes is set to $0$.
A span node $u_{l,m}$ represents a partition consisting of token nodes $u_{0,2^{l}*m+1},\cdots,u_{0,2^{l}*(m+1)}$.

To reduce the distance of information transmission, we do not directly use the tree structure to construct the edges in graph since the path is long for two long-distance tokens in the tree structure.

%The internal nodes then correspond to spans covering the words mapped to the descendant leaves.

We construct two kinds of edges:

\paragraph{Affiliated Edges}

Given a span node $u_{l,m}$, we add a directed edge from each of its contained token nodes $u_{0,2^{l}*m+1},\cdots,u_{0,2^{l}*(m+1)}$. There are $2^{l}$ edges $u_{0,2^{l}*m+i} \rightarrow u_{l,m} (1\leq i\leq 2^l)$.

The role of affiliated edges is to shorten the path between a span node and its corresponding token nodes.
With the affiliated edges, the representation of a span node is computed by directly aggregating the information from its contained token nodes.

Although we do not adopt the tree structure, the affiliated edges still can incorporate the inductive bias of the hierarchical linguistic structure within the sentence.

\paragraph{Contextual Edges}

The power of Transformer comes from relating every pair of tokens. To reduce the computation complexity while retaining the ability to capture long-range context, we model the context with a fine-to-coarse strategy. For a leaf node, to model its local context, we connect it to neighbor token nodes or lower-level span nodes. Similarly, we connect it to higher-level span nodes for long-range context.

%To model the local context of a leaf node, we need connect it to the lower-level span nodes (or token nodes) to capture the detailed information. For the long-range context, we just connect it to the higher-level span nodes.

In detail, for a leaf node $u_{0,i}$, we add the incoming edges from the different granularity. For simplicity, we describe the process of constructing edges from its right context of node $u_{0,i}$. The edges from the left context is conducted similarly.

We use a hyper-parameter $k$ to determine the connection density of the graph. We add $k$ edges per level to capture the information from the right context.

For node $u_{0,i}$, its contextual nodes are
\begin{align}
 u_{0,p_0},& \cdots,u_{0,p_0+k-1},\\
  u_{1,p_1},&\cdots,u_{1,p_1+k-1},\\
  & \cdots\\
 u_{l,p_l},&\cdots,u_{1,p_l+k-1},\\
 &\cdots,
\end{align}
where $p_l$ is the start index at level $l$ and can be computed recursively: $p_l=\mathrm{parent}(p_{l-1}+k)$ and $p_0=i+1$.

For the sake of computation efficiency, when the index $p_l+k-1$ is odd, we also add its next node in the same layer as the contextual nodes. Thus, the start index at next level is $p_{l+1}=\mathrm{parent}(p_{l}+k+1)$.

In practice, it is easy to find the contextual nodes in a recursive fashion.
Given a leaf node $u$, the whole procedure is described in Algorithm~\ref{algo:topdown}.

\begin{algorithm}[!ht]
\begin{algorithmic}
\caption{Finding contextual nodes}
\label{algo:topdown}
\Function{Neighbors}{$u, k$}
\State $\mathcal{N} \gets \{u\}, l \gets \mathrm{left}(u), r \gets \mathrm{right}(u)$
\Repeat
\For{$i \gets 1$ \textbf{to} $k$}
\State $\mathcal{N} \gets \mathcal{N} \cup \{l, r\}$
\State $l \gets \mathrm{left}(l)$

\State $r \gets \mathrm{right}(r)$

\EndFor
\State $l \gets \mathrm{parent}(l)$
\State $r \gets \mathrm{parent}(r)$
\Until $l$ and $r$ reach the boundary
\State \Return $\mathcal{N}$
\EndFunction
\end{algorithmic}
\end{algorithm}

After collect all the contextual nodes, we add a directed edge from each contextual node to node $u_{0,i}$.

Finally, for a sequence with length $n$,
we can construct a directed graph $\mathcal{G}$. The number of nodes is $O(2n)$, the number of edges is $O(kn\log n/k)$.
We can see that the distances between any two token nodes are no greater than 2 in graph $\mathcal{G}$.
This property enables our model to learn long-term dependencies easily. %: a claim proposed by \citet{vaswani2017attention}.

\subsection{Graph Update}

After graph $\mathcal{G}$ being constructed, we update representations of all nodes via Graph Self-Attention (GSA) described in Section \ref{sec:trans-gnn}.

Since $\mathcal{G}$ is a directed graph, for a given node $u$, its neighbours $\mathcal{A}(u)$ is set to all its predecessor nodes in $\mathcal{G}$.
If we set $\mathcal{A}(u)$  to all the token nodes, we recover the model to the vanilla Transformer.

Recall that the predecessors of a token node is the multi-scale spans it attending to, while the predecessors of a span node are all its contained token nodes, as illustrated in Figure \ref{fig:tree}.
Therefore, BPT connected each two tokens via at most two edges.

In our experiments, we update all nodes synchronously within the graph layer.
The representations of span nodes are initialized with all zeroes, while the representations of token nodes are initialized with the corresponding word embeddings.

We can stack multiple graph layers as in vanilla Transformer, where each layer gets its own $\mathbf{W}_\cdot^\cdot$ and $\mathbf{W}_O$.
Algorithm~\ref{algo:pseudo} demonstrates the overall update algorithm.

\begin{algorithm}[!htb]
\caption{The update of graph}
\label{algo:pseudo}
\begin{algorithmic}[1]
\Require {$\mathcal{G=(V,E)}$ the underlying graph, $N$ the number of layers, $\mathbf{H}^{0}$ initial hidden states}
\For {$i$ := $1 $ to $ N$}:
\State $\mathbf{Z}^i \gets \textrm{norm}\left(\mathbf{H}^{i-1} + \textrm{GSA}^{(i)}\left(\mathcal{G}, \mathbf{H}^{i-1}\right)\right)$
\State $\mathbf{H}^i \gets \textrm{norm}\left(\mathbf{Z}^i + \textrm{FFN}^{(i)}\left(\mathbf{Z}^i\right)\right) $
\EndFor
\State \Return ${\mathbf{H}}^N$
\end{algorithmic}
\end{algorithm}

Depending on the downstream tasks, we either take as output of representation of the root node in the final layer (e.g. in text classification and natural language inference), or the representations of all the token nodes in the final layer (e.g. in language modeling and machine translation).

\subsection{Relative Positional Encoding on Tree}

As in \cite{shaw2018self}, introducing the relative distances between words in computing the self-attention helps encode the relative order among tokens.
Here we draw a similar analogy on the tree.
For each node $v$ in $\mathcal{A}(u)$, we consider the relative positional difference on the tree between $u$ and $v$, and assign a latent representation $r_{v,u}$ of such difference:

\begin{itemize}
    \item $r_{v, u} = r^{\text{self}}$ if $v = u$.
    \item $r_{v, u} = r^{\text{left}}_{j, i}$ or $r^{\text{right}}_{j, i}$, if $v$ is the $i$-th left/right node to join the neighborhood set of $u$ at the $j$-th level in Algorithm~\ref{algo:topdown} of finding top-down context nodes.
    \item $r_{v, u} = r^{\text{anc}}_j$, if $u$ is the ancestor of $v$ in the tree at level $j$.

\end{itemize}

All the $r^{\text{self}}$, $r^{\text{left}}_{j,i}$, $r^{\text{right}}_{j,i}$ and $r^{\text{anc}}_j$  are trainable parameters.

Then, we modify Eq.~\eqref{eqn:no-relpos} to include positional representations:
\begin{equation}\small
\begin{gathered}
    \mathbf{R}^u = \text{concat}(\{r_{v,u} \mid v \in \mathcal{A}(u)\}), \\
    \mathrm{head}_i^u = \text{softmax}\left(\dfrac{\mathbf{Q}^u_i \left(\mathbf{K}^u_i +  \mathbf{R}^u \right)^T}{\sqrt{d}}\right) \mathbf{V}^u_i.
\end{gathered}
\end{equation}
Note that the relative positional representations are shared across attention heads, which is the same as in \cite{shaw2018self}, and each layer gets its own set of positional representations.

When $k$ is set to be larger then the sentence length, our model degenerates to Vanilla Transformer with positional encodings. In the following section we will show that a small $k$ (e.g. 4) is enough for achieving good performance in word level NLP tasks.

\section{Experiments}

%Our experiments focus on using the root for sentence level tasks (text classification and natural language inference) and tokens for sequence level tasks (language model, and translation).

We measure the performance of BPT on variety of tasks at both sentence level and document level. On document level tasks, we achieved state-of-the-art performance on language modeling, machine translation and text classification. For sentence level tasks, BPT performs consistently better then vanilla Transformer and Star Transformer, suggesting the inductive bias encoded by BPT is reasonable and effective for natural language.
The experimental results show the superior ability of BPT in modeling the long-range context.

\subsection{Text Classification}

We use SST-5 dataset \citep{socher2013recursive} and IMDB dataset \citep{maas2011learning} to measure the performance of our model on classification for short and long text.  The former has fine-grained labels with 215,154 phrases in 11,855 sentences with average length of 19, and the latter has positive/negative labels on 50,000 multi-sentence reviews with average length 294.
We use pre-trained GloVe embedding \citep{pennington2014glove} as input features and fixed them during training.
The hidden size of all our models are set to 300.
For IMDB, we apply the same training/validation set split ratio (0.9) as in \citet{mccann2017learned}.

\begin{table}[!htb]
\small\setlength{\tabcolsep}{2pt}
\centering
\begin{tabular}{p{14em}cc}
\toprule
Model                          & SST-5 & IMDB                  \\ \midrule
\bf BPT & 52.71(0.32)       & \bf 92.12(0.11) \\
Star Transformer         & 52.9 & 90.50 \\
Transformer              & 50.4 & 89.24 \\
Bi-LSTM \citep{li2015tree}                  & 49.8 & - \\
Tree-LSTM \citep{socher2013recursive}                & 51.0 & -\\
QRNN \cite{DBLP:conf/iclr/0002MXS17} & - & 91.4  \\ \midrule
BCN+Char+CoVe \cite{mccann2017learned}      & 53.7 & 91.8          \\
\bottomrule
\end{tabular}
\caption{Test accuracy on SST-5 and IMDB. In BPT, $k=2$ and $k=4$ for SST and IMDB respectively. The last model used word embeddings pretrained with translation and additional character-level embeddings. }
\label{tbl:classification}
\end{table}

We report the average test accuracy of BPT of 10 runs in Table \ref{tbl:classification}, the value inside brackets indicates standard derivation. On SST-5, our model outperforms Transformer and LSTM based models. On IMDB, our proposed model outperforms a bidirectional LSTM initialized with pre-trained character embedding and CoVe embedding \cite{mccann2017learned}.

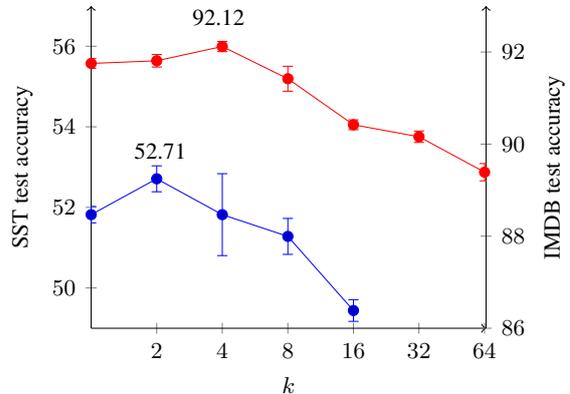
\begin{figure}[!htb]
    \centering
\resizebox {1.0\linewidth} {!} {
\begin{tikzpicture}[font=\small]
\begin{axis}[
    xmin=0,
    xmax=65,
    ymin=49,
    ymax=57,
    xmode=log,
    log basis x=2,
    xticklabel=\pgfmathparse{2^\tick}\pgfmathprintnumber{\pgfmathresult},
    %log ticks with fixed point,
    %xtick={0,100,...,500},
    axis lines=middle,
    axis line style={->},
    ylabel near ticks,
    xlabel near ticks,
    xlabel={$k$},
    ylabel={SST test accuracy},
    axis y line*=left]
\addplot+[error bars/.cd,y dir=both,y explicit]
coordinates {
    (1,51.82) +- (0, 0.204)
    (2,52.71) +- (0, 0.322)
    (4,51.82) +- (0, 1.017)
    (8,51.28) +- (0, 0.446)
    (16,49.44) +- (0, 0.269)
};\label{plot_one}
\end{axis}
\begin{axis}[
    xmin=0,
    xmax=65,
    ymin=86,
    ymax=93,
    hide x axis,
    xmode=log,
    %log ticks with fixed point,
    %xtick={0,100,...,500},
    axis lines=middle,
    axis line style={->},
    xlabel={$k$},
    ylabel={IMDB test accuracy},
    axis y line*=right,
    ylabel near ticks,
    xlabel near ticks,]
\addplot+[red, mark options={fill=red}] [error bars/.cd,y dir=both,y explicit, error bar style={red}]
coordinates {
    (1,91.752) +- (0, 0.105)
    (2,91.81) +- (0, 0.136)
    (4,92.12) +- (0, 0.111)
    (8,91.42) +- (0, 0.271)
    (16,90.42) +- (0, 0.107)
    (32,90.16) +- (0, 0.124)
    (64,89.39) +- (0, 0.187)
};
%\addlegendimage{/pgfplots/refstyle=plot_one}\addlegendentry{SST test accuracy}
%\addlegendentry{IMDB test accuracy}
\end{axis}
\node[rectangle] (A) at (1.75,4.3) {92.12};
\node[rectangle] (A) at (0.95,2.45) {52.71};
\end{tikzpicture}
}
    \caption{Effects of hyperparameter $k$.}
    \label{fig:bpt-cls}
\end{figure}

On IMDB, our model outperforms Vanilla Transformer and Star Transformer by a large margin: 1.62 and 2.88 respectively. To study the effect of $k$ on final accuracy, we tried different $k\in\{1,2,4,8,16,32,64\}$. Figure \ref{fig:bpt-cls} shows a large $k$ does not bring benefits, though it increases the graph density and time/memory cost of BPT. The best performance was obtained at $k=2$ and $k=4$ for SST and IMDB respectively, which is a small value.

\subsubsection{Sensitivity to Sequence Shift}
Since BPT divide sequence in binary fashion, a concern is whether a shift in sequence affects its performance. To measure if the output of BPT is sensitive to shift, we take the model trained on SST with best validation loss and evaluate it in a setting different from training: we append $n$ placeholder symbols in the front of each sentence, and initialize their embedding with all zeros. We varies $n$ from $0$ to $7$ and found out the test accuracy changes very little as shown in Table \ref{tbl:shift}, suggesting our model is robust towards shift.

\begin{table}[!htb]\small\setlength{\tabcolsep}{3pt}
\centering
\begin{tabular}{cl|cl}
\toprule
Shift Offset& Test Accuracy & Shift Offset& Test Accuracy\\
\midrule
0     & 52.71(0.32) & 4 & 52.18(0.22)  \\
1     & 52.50(0.29) & 5 &  51.90(0.16)  \\
2     & 52.81(0.18) & 6 & 51.85(0.35)  \\
3     & 52.56(0.22) & 7 & 51.55(0.29)  \\
\bottomrule
\end{tabular}
\caption{Accuracy with different sequence shift on SST-5.}
\label{tbl:shift}
\end{table}

\begin{comment}
\begin{table}[!htb]
\centering
\begin{tabular}{ll}
\toprule
shift & test accuracy \\
\midrule
0     & 52.71(0.32)   \\
1     & 52.50(0.29)   \\
2     & 52.81(0.18)   \\
3     & 52.56(0.22)   \\
4     & 52.18(0.22)   \\
5     & 51.90(0.16)   \\
6     & 51.85(0.35)   \\
\bottomrule
\end{tabular}
\end{table}
\end{comment}

\subsection{Language Modeling}
\label{sec:char}

To see how BPT exploits with long-term dependencies, we evaluate our model on Character Level Language Modeling datasets of moderate size: Enwiki8~\cite{mmllc:2009} and Text8~\cite{mmllc:2009}. We use bits-per-character(bpc for short, the lower the better) to measure the performance of our model.

Character level tasks require more fine-grained interactions between characters, we select a much larger $k=64$ for such tasks. The baseline models we select are multi-scale RNN based models \cite{ChungAB17, ZillySKS17, krause2016multiplicative} and Transformer-based models \cite{al2018character, dai2019transformer, sukhbaatar2019adaptive}. All Transformers use the same base setting ($12$ layers, $d=512$, $d_{ff}=2048$) for fair comparison.

\begin{table}[!htb]
\small\setlength{\tabcolsep}{2pt}
\centering
\begin{tabular}{p{14em}ccc}
\toprule
Model              & Enwiki8 & Text8 & Params \\
\midrule
HM-LSTM~\cite{ChungAB17}           & -      & 1.29 & 35M  \\
Recurrent Highway~\cite{ZillySKS17}  & -      & 1.27  & 45M \\
mLSTM~\cite{krause2016multiplicative}             & 1.24   & 1.27 & 45M \\
\midrule
Transformer~\cite{al2018character} & 1.11   & 1.18 & 44M \\
Transformer-XL~\cite{dai2019transformer}     & 1.06   & - & 41M \\
Adaptive Span~\cite{sukhbaatar2019adaptive} & 1.02 & 1.11 & 39M \\
\midrule
BPT ($k=64$, $l=8192$)       & \bf{1.02} & \bf{1.11} & \bf{38M} \\
\bottomrule
\end{tabular}
\caption{Test BPC on Enwiki8/Text8.  Note that Transformer-XL can be only used for language modeling. $l$ denotes the context length.}
\label{tbl:char}
\end{table}

In Table \ref{tbl:char}, we show that BPT can achieve state-of-the-art performance on both datasets with a small number of parameters.

To compare different sparse attention patterns, we fix the context length: $l=512$ and see how the performance of different models varies as we change the \textit{attention degree} (the number of incoming edges of each token in the context of viewing Transformer as Graph Neural Networks). For BPT, we select different $k\in \{1,2,4,8,16,32,64,128\}$, for Sparse Transformer \citep{child2019sparsetransformer}, we use the default setting described in the paper ($c=8$, $\mathrm{stride}=128$); for Restricted Transformer \citep{vaswani2017attention} (restrict self-attention to a neighborhood window of size $w$), we select $w\in\{32, 64, 128, 256, 512\}$.

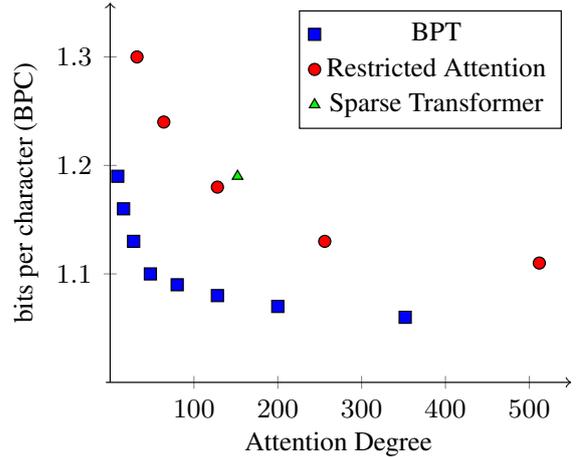
\begin{figure}[!htb]
    \centering
\resizebox {1.0\linewidth} {!} {
\begin{tikzpicture}[font=\small]
\begin{axis}[
    xmin=-5,
    xmax=550,
    ymin=1.0,
    ymax=1.35,
    xtick={0,100,...,500},
    axis lines=middle,
    axis line style={->},
    x label style={at={(axis description cs:0.5,-0.1)},anchor=north},
    y label style={at={(axis description cs:-0.12,.5)},rotate=90,anchor=south},
    xlabel={Attention Degree},
    ylabel={bits per character (BPC)}]
\addplot[only marks,mark=square*,mark options={fill=blue}] table [x=a, y=b, col sep=comma] {plot-char-1.data};
\addplot [only marks,mark options={fill=red}] table [x=a, y=b, col sep=comma] {plot-char-2.data};
\addplot [only marks,mark=triangle*,mark options={fill=green}] table [x=a, y=b, col sep=comma] {plot-char-3.data};
\legend{BPT,Restricted Attention,Sparse Transformer}
\end{axis}
\end{tikzpicture}
}
    \caption{Test BPC on Enwiki8 with different $k$.}
    \label{fig:bpt-deg}
\end{figure}

Figure \ref{fig:bpt-deg} suggests that BPT's fine-to-coarse sparse attention is more effective than Restricted Transformer and Sparse Transformer: with the same attention degree, BPT always gets better performance.

To see how BPT exploits long-term dependency, we fixed $k$ to $64$ and varies context length in $\{512, 1024, 2048, 4096, 8192\}$. We do not try context length longer than $8192$ because its exceeds the average article length in Enwik8 and Text8. As shown in Table \ref{tbl:lm-contextlength}, the performance increases with the context length.

\begin{table}[!htb]
\centering\small
\begin{tabular}{lll}
\toprule
Context Length & Enwik8 & Text8 \\
\midrule
512                                                      & 1.07   & 1.16  \\
1024                                                     & 1.05   & 1.14  \\
2048                                                     & 1.03   & 1.13  \\
4096                                                     & 1.02   & 1.12  \\
8192                                                     & 1.02   & 1.11 \\
\bottomrule
\end{tabular}
\caption{Test BPC on Enwiki8/Text8 with different context lengths.}
\label{tbl:lm-contextlength}
\end{table}

\subsection{Machine Translation}

BPT can also be applied to Encoder-Decoder frameworks by replacing backbone network in \citet{vaswani2017attention} from Transformer to BPT.

In this section we evaluate two settings: Document-Level and Sentence-Level Machine Translation. In Document-Level Machine Translation tasks, the self-attention in both encoder and decoder are applied at document level, while the attention between encoder and decoder are applied between aligned sentences. For a mini-batch of sentence pairs with source sentences of lengths $\{n_i\}$ and target sentences of lengths $\{m_i\}$, the number of connections are $\sum_{i} k n_i \log (\sum_{i} n_i / k)$ for encoder, $\sum_{i}k m_i \log (\sum_{i} m_i / k) $ for decoder, and $\sum_{i} n_i\cdot m_i$ for attention between encoder and decoder.

\subsubsection{Document Level Machine Translation}

We conduct experiments with TED Talks Chinese-to-English(Zh-En) dataset from IWSLT 2014 and 2015 \citep{cettolo2012wit3, cettolo2016iwslt}, the average document length is 120 (in sentences). For each sentence, we take its preceding context of fixed length, and their corresponding translations as a single sample.

The baseline models are HAN-NMT \cite{miculicich-etal-2018-document} and  Transformer+cache \cite{tu2018learning}. We follow the setting of \citet{miculicich-etal-2018-document} with a vocabulary size of 30k for both Chinese and English, and use \textit{dev2010} for development and \textit{tst2010-2013} for testing. Unlike previous models, our model is trained from scratch and do not require pre-training on sentence-level translation tasks.

\begin{table}[!htb]
\small
\centering
\begin{tabular}{ll}
\toprule
Model                         & BLEU  \\
\midrule
Transformer \cite{vaswani2017attention} & 16.87 \\
Transformer+cache \cite{tu2018learning} & 17.32 \\
HAN-NMT \cite{miculicich-etal-2018-document} & 17.78 \\
\midrule
Transformer (ours, single sentence) & 18.91 \\
BPT ($k=4$, single sentence) & 19.19 \\
BPT ($k=4, l=64$) & 19.84 \\
\bottomrule
\end{tabular}
\caption{BLEU score on IWSLT 2015 Zh-En}
\label{tbl:doc-mt}
\end{table}

In Table \ref{tbl:doc-mt} we show that with careful selection of hyper-parameters, Transformer trained at sentence-level could beat reported results of previous Document-Level models. BPT with $k=4$ and context length of $32$ could further improve the baseline result by $0.93$ in terms of BLEU score, which is a significant margin.

We also examine the effect of context length and $k$ on final BLEU scores, the results are shown in Table \ref{tbl:doc-mt-ctx}.  Similar to  \citet{tu2018learning} and  \citet{miculicich-etal-2018-document}, we found a small context length is enough for achieving good performance on IWSLT for Document-Level Translation. However, as we increases context size, the performance of BPT does not get worse as these models and Transformers, suggesting the inductive bias encoded by BPT makes the model less likely to overfit.

\begin{table}[!htb]
\small
\centering
\begin{tabular}{lllll}
\toprule
Context length & 0     & 32    & 64    & 128   \\
\midrule
Transformer    & 18.85 & 18.66 & 17.59 & 15.55 \\
BPT (k=4)       & 19.19 & 19.84 & 19.71 & 19.84 \\
BPT (k=8)       & 19.13 & 19.59 & 19.78 & 19.60 \\
\bottomrule
\end{tabular}
\caption{BLEU score vs context length on different models}
\label{tbl:doc-mt-ctx}
\end{table}

\subsubsection{Sentence Level Machine Translation}

IWSLT is a relatively small dataset with 0.21M sentence pairs, to see if BPT scales to large dataset, we train a BPT on WMT14 English-to-German dataset with 4.5M sentence pairs.

We follow the same setting as \cite{vaswani2017attention}, but to replace the Transformer encoder/decoder with a BPT encoder/decoder.
The number of parameters remains the same.
The baseline model we select is Transformer(base).
We trained the network for 40 epochs and take the average of last 10 checkpoint for decoding, the beam size is set to $5$.

\begin{table}[!htb]
\small
\centering
\begin{tabular}{ll}
\toprule
Model                         & BLEU  \\
\midrule
ByteNet~\cite{kalchbrenner2016neural}                       & 23.75 \\
GNMT+RL~\cite{wu2016google}                       & 24.6  \\
ConvS2S~\cite{GehringAGYD17}                     & 25.16 \\
Transformer~\cite{vaswani2017attention}       & 27.3  \\
\midrule
Transformer (our implementation) & 27.2 \\
BPT ($k=1$) & 26.9 \\
BPT ($k=2$) & 27.4 \\
BPT ($k=4$) & \textbf{27.6} \\
BPT ($k=8$) & 26.7 \\
\bottomrule
\end{tabular}
\caption{BLEU score on newstest 2014}
\label{tbl:wmt}
\end{table}

Table \ref{tbl:wmt} report the de-tokenized SacreBLEU score \footnote{Setting: \lstinline[basicstyle=\ttfamily]{BLEU+c.mixed+l.en-de+#.1+s.exp+t}
\lstinline[basicstyle=\ttfamily]{.wmt14+tok.intl+v.1.4.1}} \cite{post-2018-call} of BPT and Vanilla Transformer on test set: newstest 2014. In the setting of $k=2$ and $k=4$, BPT outperforms Vanilla Transformer with the same number of parameters and a sparse attention pattern.

The best setting of BPT on WMT14 is $k=4$, the same as the best setting of BPT on Document-Level Machine Translation(IWSLT) and Text Classification(IMDB), suggesting $k=4$ a general setting for word-level NLP tasks, on both small and large datasets.

\subsection{Throughput and GPU Memory Footprint}

\label{sec:mem}

BPT improves the time/space complexity of Transformer models from $O(d\cdot n^2)$ to $O(d\cdot k\cdot n\log n/k)$ in theory, such speedup cannot be achieved by tensor-based attention operators. To address this problem, we designed a set of CUDA kernels for sparse attentions\footnote{the speed of BPT could be further improved with better optimized kernels}.

\begin{figure}[!htb]
\resizebox {0.9\linewidth} {!} {
\begin{tikzpicture}[font=\small]
\begin{axis}[
    xmin=0,
    xmax=9000,
    ymin=18000,
    ymax=140000,
    axis lines=middle,
    axis line style={->},
    ylabel near ticks,
    xlabel near ticks,
    xlabel={length of sequence},
    ylabel={Throughput (tokens/s)},
    legend style={at={(1,1.1)},anchor=north east},
]
\addplot+[mark size=1.2] table [x=a, y=c, col sep=comma] {plot-speed-1.data};
\addplot+[mark size=1.2]table [x=a, y=c, col sep=comma] {plot-speed-2.data};
\addplot+[mark size=1.2] table [x=a, y=e, col sep=comma] {plot-speed-2.data};
\addplot+[mark size=1.2] table [x=a, y=g, col sep=comma] {plot-speed-2.data};
\addplot+[mark size=1.2,color=green] table [x=a, y=i, col sep=comma] {plot-speed-2.data};
\legend{Transformer,BPT($k=1$),BPT($k=4$),BPT($k=16$),BPT($k=64$)}
\end{axis}
\end{tikzpicture}
}
    \caption{Throughput vs sequence length}
    \label{fig:bpt-spd}
\end{figure}
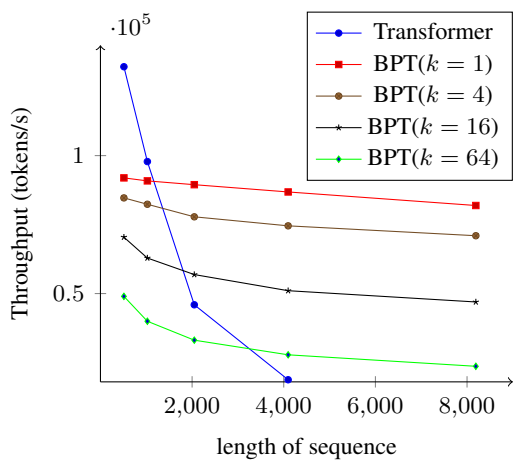

We compare the GPU memory footprint and throughput of BPT and vanilla Transformer during inference under the same setting\footnote{$N=6$, $d=512$, $d_{ff}=2048$, $h=8$} for language modeling. The $k$ is set to $1, 4, 16, 64$ respectively, covering best settings for word-based tasks($k=4$) and character-based tasks($k=64$). We fix the number of tokens to 8192 each batch and varies the sequence length. Figure \ref{fig:bpt-mem} and \ref{fig:bpt-spd} depicts how the GPU memory and speed varies as we increases sequence length.

\begin{figure}[!htb]
\resizebox {0.9\linewidth} {!} {
\begin{tikzpicture}[x=0.7cm,y=0.5cm,font=\small]
\begin{axis}[
    xmin=0,
    xmax=9000,
    ymin=1.5,
    ymax=11,
    axis lines=middle,
    axis line style={->},
    ylabel near ticks,
    xlabel near ticks,
    xlabel={length of sequence},
    ylabel={GPU Memory(GB)}]
\addplot+[mark size=1.2] table [x=a, y=b, col sep=comma] {plot-speed-1.data};
\addplot+[mark size=1.2]table [x=a, y=b, col sep=comma] {plot-speed-2.data};
\addplot+[mark size=1.2] table [x=a, y=d, col sep=comma] {plot-speed-2.data};
\addplot+[mark size=1.2] table [x=a, y=f, col sep=comma] {plot-speed-2.data};
\addplot+[mark size=1.2,color=green] table [x=a, y=h, col sep=comma] {plot-speed-2.data};
\legend{Transformer,BPT($k=1$),BPT($k=4$),BPT($k=16$),BPT($k=64$)}
\end{axis}
\end{tikzpicture}
}
    \caption{GPU memory cost vs sequence lengh}
    \label{fig:bpt-mem}
\end{figure}
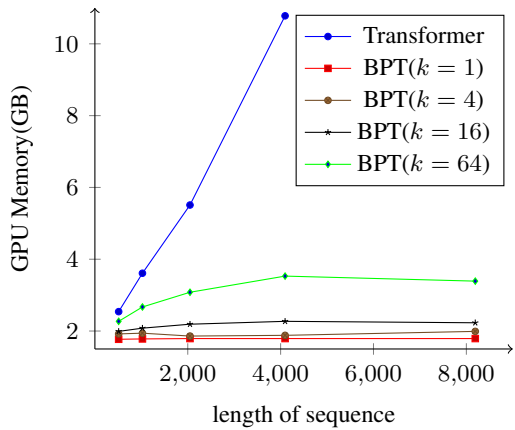

We show that BPT consistently utilizes less GPU memory compared to Transformer, making it possible to be applied on tasks that require long sequence modeling such as time-series prediction.

As for speed, BPT increases the number of nodes from $n$ to $2n$ which brings additional overhead linear to sequence length, rendering BPT not as fast as Transformer when dealing with short text. However, as the sequence length grows, the speed of BPT is steady while Transformer become too slow for practical use.

\section{Conclusion and Future Work}

This paper introduces a hierarchical fine-to-coarse self-attention based model that is versatile and flexible for a variety of natural language processing tasks.
%Its core idea is to leverage a latent tree over different spans of spans so as to extract hierarchical features.
By imposing structural inductive bias this way we are able to strike a balance between the power of the model and training/computational efficiency.
%, arriving at an $O(n \log n)$ architecture.

%We note that viewing a model from a graph perspective does not mean that the architecture needs to be rigid. By extending local connectivity window we can modify network density per task, for instance character-level language model requires a larger local context window than word-level, as we proved experimentally.

This work can be extended in a number of interesting ways. The representations have not yet naturally captured syntactic and semantic meanings. Instead of only using the root and the token representations, other intermediate representations can be more directly exposed.
%Finally, the model can be trained in a multi-task setting, just as in BERT~\citep{devlin2018bert}.

\bibliography{acl2019}
\bibliographystyle{acl_natbib}

\appendix
\section{Appendix}
\label{sec:appendix}

\subsection{Implementation Details}
\label{app:hyper}
We use Deep Graph Library \cite{wang2019dgl} for building Binary Partition graphs.

The following table summarizes the hyper-parameters used in BPT.

\begin{table}[!htb]
\begin{tabular}{ll}
\toprule
notation & meaning                                          \\
\midrule
$B_{tok}$          & number of tokens in a batch                      \\
$B_{sent}$   & number of sentences in a batch \\
$N$          & number of (encoder) layers.                  \\
$M$          & number of (decoder) layers.                  \\
$h$          & number of heads.                                 \\
$k$          & connection density in BPT                        \\
$d_{emb}$    & embedding size                                   \\
$d$          & hidden size of the model                         \\
$d_{ff}$     & filter size in FFN sublayer                      \\
$p_i$        & dropout rate on embedding layer                  \\
$p_h$        & dropout rate on hidden layers                    \\
$p_a$        & dropout rate on attention weight                 \\
$p_c$        & dropout rate before classifier \\
avg          & model average checkpoints              \\
steps & training steps                           \\
epochs & training epochs \\
\bottomrule
\end{tabular}
\end{table}

\begin{table}[!htb]
\centering
\begin{tabular}{lll}
\toprule
          & SST & IMDB \\
\midrule
$N$       & 4   & 5    \\
$d_{emb}$ & 300 & 300  \\
$d$       & 300 & 300  \\
$d_{ff}$  & 600 & 600  \\
$h$       & 6   & 6    \\
$p_i$     & 0.4 & 0.5  \\
$p_h$     & 0.1 & 0.1  \\
$p_a$     & 0.3 & 0.3  \\
$p_c$     & 0.4 & 0.5  \\
$B_{sent}$       & 1024 & 32  \\
epochs    & 40  & 40   \\
\bottomrule
\end{tabular}
\caption{Hyper-parameters for Text Classification}
\label{tbl:hyper-tc}
\end{table}

\begin{table}[!htb]
\centering
\begin{tabular}{lll}
\toprule
          & enwik8 & text8 \\
\midrule
$N$       & 12   & 12    \\
$d_{emb}$ & 512 & 512  \\
$d$       & 512 & 512  \\
$d_{ff}$  & 2048 & 2048  \\
$h$       & 8   & 8    \\
$p_i$     & 0.1 & 0.1  \\
$p_h$     & 0.1 & 0.1  \\
$p_a$     & 0.1 & 0.1  \\
$p_c$     & 0.1 & 0.1  \\
$B_{tok}$       & 32768 & 32768 \\
steps     & 400000 & 600000   \\
\bottomrule
\end{tabular}
\caption{Hyper-parameters for Language Modeling}
\label{tbl:hyper-lm}
\end{table}
\begin{table}[!htb]
\centering
\begin{tabular}{lll}
\toprule
          & IWSLT & WMT \\
\midrule
$N$       & 6   & 6    \\
$M$       & 6   & 6    \\
$d_{emb}$ & 512 & 512  \\
$d$       & 512 & 512  \\
$d_{ff}$  & 2048 & 2048  \\
$h$       & 8   & 8    \\
$p_i$     & 0.1 & 0.1  \\
$p_h$     & 0.1 & 0.1  \\
$p_a$     & 0.1 & 0.1  \\
$p_c$     & 0.1 & 0.1  \\
$B_{sent}$       & $128$ & $1024$ \\
avg       & 10 & 10   \\
epochs    & 40 & 40   \\
\bottomrule
\end{tabular}
\caption{Hyper-parameters for Machine Translation}
\label{tbl:hyper-mt}
\end{table}

Table \ref{tbl:hyper-tc}, \ref{tbl:hyper-lm} and \ref{tbl:hyper-mt} lists the hyper-parameters we use in Text Classification, language Modeling and Machine Translation respectively.

For full details please refer to the configurations in our source code: \url{https://github.com/yzh119/BPT/tree/master/configs}.
\end{document}